\documentclass[10pt,letterpaper]{article}
\usepackage[top=0.85in,left=2.75in,footskip=0.75in,marginparwidth=2in]{geometry}

\usepackage[utf8]{inputenc}

\usepackage{cite}

\usepackage{algorithm2e}

\usepackage{nameref,hyperref}

\usepackage[right]{lineno}

\usepackage{microtype}
\DisableLigatures[f]{encoding = *, family = * }

\raggedright
\usepackage{changepage}

\usepackage[aboveskip=1pt,labelfont=bf,labelsep=period,singlelinecheck=off]{caption}

\makeatletter
\renewcommand{\@biblabel}[1]{\quad#1.}
\makeatother

\usepackage{lastpage,fancyhdr,graphicx}
\usepackage{epstopdf}
\fancyheadoffset[L]{2.25in}
\fancyfootoffset[L]{2.25in}

\usepackage{color}

\definecolor{Gray}{gray}{.25}

\usepackage{graphicx}
\usepackage{multirow}
\usepackage{amsmath}
\usepackage{amssymb}

\usepackage{sidecap}

\usepackage{wrapfig}
\usepackage[pscoord]{eso-pic}
\usepackage[fulladjust]{marginnote}
\reversemarginpar

\begin{document}
\vspace*{0.35in}


\begin{flushleft}
{\Large
\textbf\newline{Sequence-based Sleep Stage Classification using \hspace{2.0cm} Conditional Neural Fields}
}
\newline
\\
Intan Nurma Yulita\textsuperscript{1,2},
Mohamad Ivan Fanany\textsuperscript{1},
Aniati Murni Arymurthy\textsuperscript{1},
\\
\bigskip
\bf{1} Machine Learning and Computer Vision Laboratory, \\Faculty of Computer Science, Universitas Indonesia\\
\bf{2} Department of Computer Science, Universitas Padjadjaran\\
\bigskip
* intanurma@gmail.com

\end{flushleft}

\providecommand{\keywords}[1]{\textbf{\textit{Keywords---}} #1}

\section*{Abstract}
Sleep signals from a polysomnographic database are sequences in nature. Commonly employed analysis and classification methods, however, ignored this fact and treated the sleep signals as non-sequence data. Treating the sleep signals as sequences, this paper compared two powerful unsupervised feature extractors and three sequence-based classifiers regarding accuracy and computational (training and testing) time after 10-folds cross-validation. The compared feature extractors are Deep Belief Networks (DBN) and Fuzzy C-Means (FCM) clustering. Whereas the compared sequence-based classifiers are Hidden Markov Models (HMM), Conditional Random Fields (CRF) and its variants, i.e., Hidden-state CRF (HCRF) and Latent-Dynamic CRF (LDCRF); and Conditional Neural Fields (CNF) and its variant (LDCNF). In this study, we use two datasets. The first dataset is an open (public) polysomnographic dataset downloadable from the Internet, while the second dataset is our polysomnographic dataset (also available for download). For the first dataset, the combination of FCM and CNF gives the highest accuracy (96.75\%) with relatively short training time (0.33 hours). For the second dataset, the combination of DBN and CRF gives the accuracy of 99.96\% but with 1.02 hours training time, whereas the combination of DBN and CNF gives slightly less accuracy (99.69\%) but also less computation time (0.89 hours). 
\bigskip

\noindent\keywords{Sleep stage, Conditional Neural Fields, Deep Belief Networks, Fuzzy C-Means Clustering, Classification}


\section*{Introduction}
\label{Introduction}

Accurate sleep stage classification is paramount in telemedicine and home care treatment of patients with a sleep disorder. Most sleep stage classification methods, however, ignored the nature of sleep signals as a sequence signal. We argue that this sequence nature is necessary to be considered and might bring more insightful pattern which ultimately delivers accurate classification. Many previous sleep stage classification methods treat sleep data as non-sequence data. They used wavelet and artificial neural networks \cite{Jai12}, shallow classifiers \cite{Gir15}, and Deep Belief Networks \cite{Zha15}. On the other hand, HMM is a widely known method for labeling sequence data \cite{Rab93}. This approach has been used in many areas such  as  speech \cite{Yul12},  gesture \cite{Elm08}, marine \cite{Spa12}, health \cite{Ben04},and Biology \cite{Byu09}, and also sleep stage \cite{Mar12} . However, HMM has some limitations that the distribution of the conditional probability for hidden variable only at one time segment. To achieve better performance, Laferty et al. proposes the use of Conditional Random Fields (CRF) to overcome the weakness of HMM \cite{Laf2}. CRF is a undirected graph model that combines the features of a complex observational sequence.  Unlike HMM, CRF also accommodates bias in HMM \cite{Qua04}.

Due to its success in the sequence labeling, CRF variants have been developed with some additional functions such as hidden-state (in HCRF) to capture the interacting features in the data \cite{Gun05}. However, the HCRF eliminated the role of interacting labels, so the LDCRF was proposed to combine these factors \cite{Boo14}, \cite{Wit15}, \cite{Zha12}.  Furthermore, the CRF and LDCRF have been further developed by adding a new layer to map the complex non-linear relationship. The new layer is called the gate, which is built into the internal processes \cite{Jia09}. Both these last methods are called as CNF and LDCNF.

L\"{a}ngkvist conducts a sleep stage classification study, et al. \cite{Mar12}. They proposed an unsupervised learning for generating features by Deep Belief Networks (DBN). While not specifically treat the sleep data as a sequence, they compared the use of DBN-HMM and DBN-only classifiers. They found that the DBN-HMM (sequential treatment) gives better results DBN-only (non-sequential treatment). In this paper, we are interested in improving their classification performance by replacing the HMM part with CRF and CNF variants in handling the sequential data for sleep stage classification. L\"{a}ngkvist et al. \cite{Mar12} also found that hand-crafted features works better than machine-generated features automatically obtained from raw data. Inline with the finding, this study also employs hand-crafted features. 

We also interested to investigate further the effectiveness of DBN as unsupervised features extractor in comparison with other features extractor such as FCM clustering. We compared many configurations regarding the accuracy and computation time. The required output of Deep Belief Networks is the probability of a segment for each class. It has similarities with the concept of cluster formation in an FCM clustering, in which, a segment is a member of all clusters with different membership degree. Based on such objective, the use of FCM clustering might be more optimal due to its concept of uncertainty, which enables no information loss in constructing new features. Therefore, this paper also proposes to examine the use of FCM clustering as the extraction of new features.

\section{2.	Materials and Methods}
This chapter describes our dataset, the used of feature extraction and classification methods, and also our experiment schemes. This research compares two unsupervised feature extraction methods: DBN and FCM; and three classification methods: HMM, CRF-variants, and CNF-variants. Each of these methods is described in the following Section. 
\subsection{Dataset}
The proposes methods are evaluated using two datasets, namely:
\begin{enumerate}
\item The first dataset came from St. Vincent's  University  Hospital and University  College  Dublin which can be downloaded  from:http//physionet.org/pn3/ucddb/. This dataset is also used by L\"{a}ngkvist et al. 2012 but this paper only test the first 10 from the available 25 data. The dataset consists of a set of recording of electroencephalography (EEG), electrooculography (EOG), and electromyography (EMG). For the classification, it detects five sleep stages: Awake, Stage 1 (S1), Stage 2 (S2), Slow wave sleep (SWS), and Rapid eye movement (REM).  
\item The second dataset was obtained through a collaboration between Mitra Keluarga Kemayoran Hospital, Jakarta. The dataset was obtained from a research collaboration between Faculty of Computer Science, Universitas Indonesia, with The Mitra Kemayoran Hospital. It was about 8 hours full night sleep of polysomnogram records from five males and females \cite{Ind12}. For the classification, it also uses five sleep stages: Awake, Stage 1(S1), Stage 2 (S2), Stage 3 (S3), and Rapid eye movement (REM). The data annotation was performed by a sleep specialist from the Mitra Keluarga Kemayoran Hospital.  
\end{enumerate}

\subsection{Conditional Random Fields}
Recognition can be performed without processing dynamic interactions between labels in a sequence. However, modeling by considering the dynamics, certainly, have an advantage in exploiting the correlation between labels. Some techniques are specifically modeled sequential data aimed at extracting such correlation, one of them is CRF \cite{Ton16}. 

\begin{figure}[h]
\includegraphics[scale = 0.7]{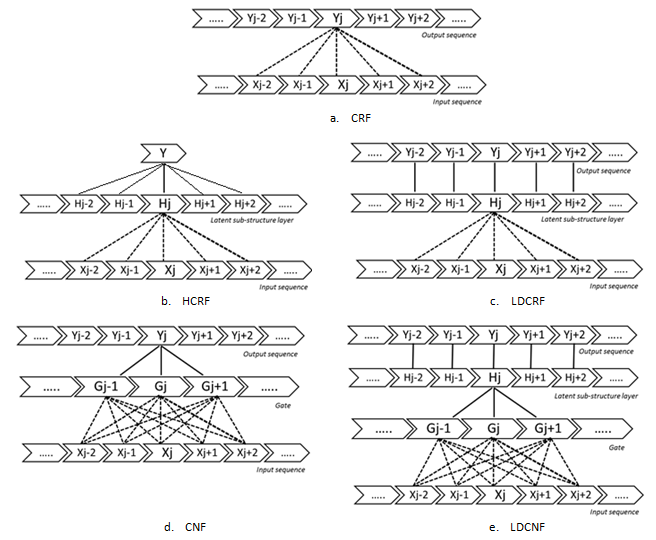}
\caption{Graphical Models}
\label{fig:models}
\end{figure}

CRF modeling is based on the joint distribution of the data sequence X and a sequence label Y\cite{Sut06}. Both these sequences have the same length.
\begin{equation}
X=x_1,x_2,x_3,x_4,…..x(n)
\end{equation}
\begin{equation}
Y=y_1,y_2,y_3,y_4,…..y(n)
\end{equation}
where $n$ is the length of the data segment, while $x(n)$ is a feature vector data from sleep, which have a sequence label $y_n$. Members of $y_n$ are the set of possible labels. In CRF, $F$ is a global feature and consists of local features that are defined as follows \cite{Sut11}:
\begin{equation}
F(Y,X)= \sum_{i}{f(Y,X,n)}
\end{equation}
Local features consist of state feature $s (y, x, n)$ or transition feature $t (y, y ', x, n)$ and each local feature vector must correspond to a weight vector $\lambda$. 
Thus, the model of CRF is defined as follows:
\begin{equation}
p(Y|X,\lambda)= \frac{\exp \lambda F(Y,X)}{Z(X)} 
\end{equation}
where
\begin{equation}
Z(X,\lambda)= \sum_{y}{P(h | \theta)P(v |h, \theta)}
\end{equation}
Training the CRF can be done through maximizing the log-likelihood of the training set. It is given as follows:
\begin{equation}
\theta(\lambda)= \sum_{k}{ \log p(Y_{k}|X_{k},\lambda)}
\end{equation}
For improving the convergence of the model, some optimization techniques that commonly used are Conjugative Gradient (CG) and Broyden-Fletcher-Goldfarb-Shanno (BFGS) algorithms. The CRF has also been developed by adding some hyperparameters such as a layer of hidden states and gates. Both are further described in the following Subsections, and also compared in Figure \ref{fig:models}.

\subsubsection{Hidden States}
CRF is limited in capturing the correlation between features in the data. To alleviate this limitation, HCRF is designed by adding an intermediate structure consisting of some hidden states into the CRF \cite{Qua07}, \cite{Ton15}.  Due to this intermediate structure, the conditional models of HCRF becomes as follows:
\begin{equation}
p(Y|X,\lambda)= \sum_{h}{p(Y|h,X.\lambda) p(h|X,\lambda)}
\end{equation}
where h is the vector of the intermediate structure. HCRF structure designed to capture the correlation between the features of the data but ignores the dynamics of intercorrelated label that has been owned previously by the CRF structure. Therefore, LDCRF was developed to combine the CRF structure and HCRF \cite{Rah16}. To address the HCRF limitations,  $p(Y|h, X,\lambda)$ of LDCRF is zero, so conditional models of LDCRF are formulated as follows:
\begin{equation}
p(Y|X,\lambda)= \sum_{h}{p(h|X,\lambda)}
\end{equation}

Using the CRF formula, it is known that:
\begin{equation}
p(Y|X,\lambda)= \frac{\exp \lambda F(Y,X)}{Z(X)}
\end{equation}
where
\begin{equation}
Z(X,\lambda)= \sum_{h}{\exp \lambda F(h,X)}
\end{equation}
\begin{equation}
F(h,X)= \sum_{i}{f(h,X,n)}
\end{equation}

\subsubsection{Gate Function}
Conditional Neural Fields (CNF) combines the advantages of CRF and Artificial Neural Networks \cite{Jia09}. CRF has the advantage in representing the structural relationship among their labels, whereas Artificial Neural Networks captures the correlation between input and output. For representing the function of artificial neural network in the CRF, a new intermediate structure is added. The new intermediate structure consists of some $K$ gate functions. These $K$ gate functions extract the input features becomes $K$ dimensions. By using the gate functions $\vartheta$ then the state feature is formulated as follows:

\begin{equation}
F(Y,X)= \sum_{i}\sum_{g=1}^{K}{\vartheta( \varphi_{g} f(Y,X,n)}
\end{equation}
As CRF, the CNF does not capture the intermediate structure inter-correlations. Thus, LDCNF was developed to overcome these deficiencies and also studied the complex non-linear relationship of input to output \cite{Lev13}.  The resulting model has two layers: gate and hidden states. Therefore, the feature state is defined as follows:
\begin{equation}
F(h,X)= \sum_{i}\sum_{g=1}^{K}{\vartheta( \varphi_{g} f(h,X,n)}
\end{equation}

\subsection{Deep Belief Networks}
Deep Belief Networks (DBN) a directed acyclic graph that is composed of some hidden variables that are arranged in layers \cite{Hin06}. DBN works in two schemes. The first scheme learns unsupervised as a feature extractor for input data to find a subset of features. The second scheme learns supervised as a classifier. The result of the first scheme becomes the input to the second scheme.

DBN is composed of several Restricted Boltzmann Machines (RBMs). The RBM has a role in getting the weights of the adjacent two layers and constructs a new feature subset. To build a model, RBM uses some visible and hidden units. The units in the visible and hidden layers are interconnected each other, but no connections between units inside visible and units inside hidden layer. The reconstruction process is done through forward and backward mechanism from the top to the bottom of the RBM and vice-versa.  Forward and backward process are often called encoding and decoding.

\subsection{FCM Clustering}
FCM clustering is a technique to group or to cluster objects into all cluster memberships \cite{Ylu13}. Although the object is in the membership of all clusters, the degree of membership for each cluster is different. It means that the higher degree of membership to a cluster, then the characteristics of object are closer to the center of this cluster. In this study, FCM used as a feature extraction so that it has the same role with DBN. The output of feature extraction to be achieved in this study is any data have a subset of features which these features have membership degree of data towards the centroids.

The FCM clustering algorithm is as follows:
\begin{enumerate}
    \item \textbf{Initialize $X$:} $X$ is the data to be processed in the FCM $n$-dimensional matrix of data and m attributes.
    \item \textbf{Determine parameter settings:} Some cluster $c$ is not altered within the processing cluster so that the number needs to be defined. Weights ($w$) controls fuzziness of the reach for each cluster.
    \item \textbf{Initialization of initial partition matrix $U$:} $U$ is an n-dimensional matrix of data and cluster $c$. This initialization determines the centroid to be set up so that if it establishes the proper initialization, then it will be more quickly to find the expected optimal value.
This matrix is defined as follows:
\begin{equation}
U_{ik}= \frac{U_{ik}}{Z_{i}}
\end{equation}
\begin{equation}
Z_{i}= \sum_{k=1}^{c}{U_{ik}}
\end{equation}
where $Z_{i}$ is used for normalization.
    \item \textbf{Calculate the cluster center 	$V_{kj}$} with $k=1,2,.. c$ and $j=1,2,.. m$: 
\begin{equation}
V_{kj}= \frac{\sum_{i=1}^n{(U_{ik})^{w} X_{ij}}}{\sum_{i=1}^n{(U_{ik})^w}}
\end{equation}

    \item \textbf{Calculate the objective function:} 
    \begin{equation}
P_{t}= \sum_{i=1}^n \sum_{k=1}^c ((\sum_{j=1}^m{(X_{ij}-V_{kj})^2})(U_{ik})^w)
\end{equation}
    
    \item \textbf{Calculate a new partition matrix:} 
    \begin{equation}
U_{ik}= \frac{(\sum_{j=1}^m{(X_{ij})^w2 X_{ij}})^\frac{-1}{w-1}}{Z_{i}}
\end{equation}
    \item \textbf{Check the stop condition:} If $(|P_{t}-P_{t-1} |<e_{t} )$ or $t>t_{max}$ then the training process has finished, but if not, $t = t + 1$ and repeat steps 4.
Where $e_{t}$  is expected e smallest and $error_{max}$ is maximum iteration

\end{enumerate}

\subsection{Design of The Proposed Method}
Sleep stage classification in this study is described in Figure \ref{fig:architecture}:
\begin{figure}[h]
\includegraphics[scale = 0.8]{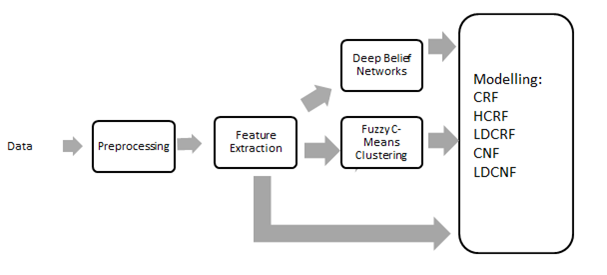}
\caption{The architecture of the proposed methods}
\label{fig:architecture}
\end{figure}
In the earlier step, all signals are preprocessed using notch filtering and down sample process. The preprocessed result is extracted to get essential characteristics.  The results of the feature extraction is treated in three scenarios before modeling the CNF, namely:
\begin{enumerate}
\item Without an additional feature extraction, such as DBN and FCM
\item With an additional feature extraction using DBN 
\item With an additional feature extraction using FCM
\end{enumerate}
Some new features are extracted by DBN as the number of classes used. In contrast with FCM Clustering, which extracts the number of clusters as a new set of features, the three scenarios have some different features.
To form a CNF model, the used window size is 0, and the maximum window size is 1000. The training for making the model convergence is done by using the Broyden-Fletcher-Goldfarb-Shanno algorithm. To test the resulted model, the experimental process is conducted through 10 cross-validation.

\section{Results and Discussion}
This Section explains the evaluation of conditional models obtained from both datasets. The evaluated parameters are the accuracy (\%) and the computation time (hours).

\subsection{St. Vincent's  Dataset}
Gate function in Conditional Neural Fields (CNF)  maps the complex nonlinear relationship of input to output. The effectiveness of this additional function in CNF is shown in Table \ref{tab:lcnf1}. The percentage of highest average accuracy was 95.98 when it used six gates, but others differed only slightly. It indicates the average accuracy of CNF tends to be stable even when changing the number of the gate. The same thing also happened on the required computation time. Without using DBN or FCM features, the CNF has achieved the average accuracy above 95 by using 28 features. However, through the implementation of DBN as feature extractor in Table  \ref{tab:lcnf2}, the CNF can reduce the computation time to within 0.04 to 0.06 hours range for each fold. Even though the accuracy is only slightly increased. Besides that, the highest accuracy of DBN achieved when used five gates. It seems that the DBN does work in giving the optimal subset of features so that the required number of gates needed to reach optimal conditions are also less.
\begin{table}[]
\small
\caption{Number of Gates in CNF for St. Vincent's  Dataset}
\label{tab:lcnf1}

\noindent\hrulefill

\smallskip\noindent

\resizebox{\columnwidth}{!}{%

\begin{tabular}{|c|c|c|c|c|c|c|c|c|c|c|c|c|c|c|}
\multirow{2}{*}{Fold} & \multicolumn{7}{c|}{Accuracy (\%)} & %
    \multicolumn{7}{c|}{Computation Time (hours)}\\
\cline{2-15}
 & g=2 & g=3 & g=4 &g=5 & g=6 & g=7 & g=8  & g=2 & g=3 & g=4 &g=5 & g=6 & g=7 & g=8\\
\hline
\hline
1	&	98.54	&	98.53	&	98.60	&	98.44	&	85.93	&	0.03	&	0.03	&	0.24	&	0.14	&	0.18	&	0.25	&	0.24	&	0.23	&	0.24	\\ \hline
2	&	89.93	&	89.96	&	89.75	&	86.22	&	89.55	&	0.02	&	0.02	&	0.21	&	0.12	&	0.14	&	0.17	&	0.17	&	0.13	&	0.09	\\ \hline
3	&	98.97	&	94.99	&	96.84	&	95.05	&	93.20	&	0.03	&	0.03	&	0.12	&	0.10	&	0.09	&	0.13	&	0.10	&	0.13	&	0.15	\\ \hline
4	&	96.87	&	97.16	&	97.06	&	97.42	&	97.38	&	0.03	&	0.03	&	0.09	&	0.08	&	0.10	&	0.10	&	0.10	&	0.11	&	0.01	\\ \hline
5	&	97.85	&	98.00	&	97.45	&	98.30	&	97.18	&	0.03	&	0.03	&	0.04	&	0.07	&	0.07	&	0.03	&	0.01	&	0.08	&	0.11	\\ \hline
6	&	82.88	&	84.09	&	93.47	&	89.71	&	81.34	&	0.02	&	0.02	&	0.11	&	0.09	&	0.12	&	0.13	&	0.09	&	0.10	&	0.07	\\ \hline
7	&	96.67	&	99.48	&	99.50	&	93.27	&	99.51	&	0.03	&	0.03	&	0.07	&	0.08	&	0.10	&	0.05	&	0.12	&	0.13	&	0.11	\\ \hline
8	&	99.61	&	98.77	&	98.79	&	99.38	&	99.98	&	0.03	&	0.03	&	0.07	&	0.06	&	0.09	&	0.02	&	0.10	&	0.10	&	0.06	\\ \hline
9	&	94.76	&	91.97	&	86.64	&	53.01	&	76.00	&	0.03	&	0.03	&	0.08	&	0.07	&	0.06	&	0.05	&	0.05	&	0.07	&	0.09	\\ \hline
10	&	99.94	&	99.94	&	99.95	&	99.94	&	99.73	&	0.03	&	0.03	&	0.07	&	0.05	&	0.06	&	0.08	&	0.05	&	0.07	&	0.06	\\ \hline
\hline																													
Average	&	95.60	&	95.29	&	95.80	&	91.07	&	91.98	&	0.03	&	0.03	&	0.11	&	0.09	&	0.10	&	0.10	&	0.10	&	0.11	&	0.10	\\ \hline
\end{tabular}
}
\end{table}
 
\begin{table}[]
\small
\caption{Number of Gates in DBN-CNF for St. Vincent's  Dataset}
\label{tab:lcnf2}
\begin{tabular}{|c|c|c|c|c|c|c|c|c|}
\hline
\multirow{2}{*}{Fold} & \multicolumn{4}{c|}{Accuracy (\%)} & %
    \multicolumn{4}{c|}{Computation Time (hours)}\\
\cline{2-9}
 & g=2 & g=3 & g=4 &g=5  & g=2 & g=3 & g=4 &g=5 \\
\hline
\hline
1	&	96.11	&	96.26	&	96.10	&	96.38	&	0.08	&	0.11	&	0.07	&	0.09	\\ \hline
2	&	88.70	&	88.68	&	88.60	&	88.64	&	0.09	&	0.05	&	0.05	&	0.04	\\ \hline
3	&	96.25	&	96.25	&	96.04	&	96.33	&	0.06	&	0.08	&	0.05	&	0.07	\\ \hline
4	&	94.78	&	94.78	&	94.41	&	97.26	&	0.05	&	0.04	&	0.05	&	0.01	\\ \hline
5	&	98.93	&	98.92	&	98.81	&	98.92	&	0.05	&	0.04	&	0.01	&	0.03	\\ \hline
6	&	93.17	&	94.06	&	93.80	&	93.17	&	0.04	&	0.05	&	0.05	&	0.04	\\ \hline
7	&	99.73	&	99.74	&	99.74	&	99.91	&	0.04	&	0.03	&	0.02	&	0.02	\\ \hline
8	&	99.70	&	99.70	&	99.70	&	99.70	&	0.04	&	0.03	&	0.03	&	0.04	\\ \hline
9	&	95.51	&	96.41	&	95.49	&	95.58	&	0.03	&	0.03	&	0.03	&	0.03	\\ \hline
10	&	99.89	&	99.89	&	99.90	&	99.89	&	0.03	&	0.04	&	0.03	&	0.03	\\ \hline
\hline																	
Average	&	96.28	&	96.47	&	96.26	&	96.58	&	0.05	&	0.05	&	0.04	&	0.04	\\ \hline

\end{tabular}
\end{table}

On the other hand, the use of FCM clustering to extract four features from 28 features seems works so that CNF gives the highest accuracy. This accuracy is even better than the use of DBN even though the different is small. Using only four features. However, the required computation time is longer since the constructing of clusters takes a long time and affect the overall execution time for the classification process. In fact, the time required is longer than the classification without using additional feature extraction. As shown in Table  \ref{tab:lcnf3}, the average computation time ranges 0.3 hours up to 0.33 hours for each fold so that the feature extraction becomes ineffective since the increased accuracy is obtained only slightly but the computation time is much larger.

\begin{table}[]
\small
\caption{Number of Gates in FCM-CNF for St. Vincent's  Dataset}
\label{tab:lcnf3}
\begin{tabular}{|c|c|c|c|c|c|c|c|c|}
\hline
\multirow{2}{*}{Fold} & \multicolumn{3}{c|}{Accuracy (\%)} & %
    \multicolumn{3}{c|}{Computation Time (hours)}\\
\cline{2-7}
 & g=2 & g=3 & g=4   & g=2 & g=3 & g=4 \\
\hline
\hline
1	&	96.08	&	97.57	&	97.91	&	0.59	&	0.60	&	0.68	\\ \hline
2	&	89.56	&	89.86	&	86.67	&	0.58	&	0.44	&	0.19	\\ \hline
3	&	98.66	&	97.09	&	97.78	&	0.34	&	0.27	&	0.38	\\ \hline
4	&	93.35	&	92.45	&	93.00	&	0.24	&	0.33	&	0.31	\\ \hline
5	&	99.30	&	98.98	&	99.05	&	0.29	&	0.29	&	0.26	\\ \hline
6	&	95.44	&	95.26	&	93.50	&	0.48	&	0.19	&	0.22	\\ \hline
7	&	98.69	&	98.78	&	99.23	&	0.14	&	0.51	&	0.29	\\ \hline
8	&	99.53	&	99.87	&	99.81	&	0.05	&	0.25	&	0.46	\\ \hline
9	&	96.41	&	97.67	&	97.78	&	0.10	&	0.12	&	0.15	\\ \hline
10	&	99.99	&	99.98	&	99.94	&	0.15	&	0.34	&	0.11	\\ \hline
 \hline													
Average	&	96.70	&	96.75	&	96.47	&	0.30	&	0.33	&	0.31	\\ \hline
\end{tabular}
\end{table}
The implementation of CNF in Table  \ref{tab:lcnf3} using four features obtained from FCM Clustering. These features represent the four clusters.  The selection of only four of these features is based on tests performed in Table  \ref{tab:lcl} which test the CNF performance for alteration of the number of clusters based on the accuracy and computation time.  The result showed that the highest accuracy was obtained when using only four clusters and the accuracy was decreased when the number of clusters increased. However, unlike the computation time, which has an ascended trend when the number of clusters increased. From the table, it is also known that the highest computation time obtained when used eight clusters. Also, if the highest accuracy is achieved when four clusters are applied even though the number of clusters is a representation of the class label for sleep stage classification which is defined five classes, then it indicates that two classes have similar characteristics. 
\begin{table}[]
\small
\caption{Number of Clusters in Fuzzy-CNF for St. Vincent's  Dataset}
\label{tab:lcl}
\begin{tabular}{|c|c|c|c|c|c|c|c|c|c|c|}
\hline
\multirow{2}{*}{Fold} & \multicolumn{5}{c|}{Accuracy (\%)} & %
    \multicolumn{5}{c|}{Computation Time (hours)}\\
\cline{2-11}
 & cl=4  &  cl=5  & cl=6  & cl=7  &  cl=8  & cl=4  &  cl=5  & cl=6  & cl=7  &  cl=8 \\
\hline
\hline
1	&	96.08	&	95.26	&	95.99	&	91.31	&	86.82	&	0.59	&	0.61	&	0.49	&	0.43	&	0.86	\\ \hline
2	&	89.56	&	91.41	&	87.49	&	88.96	&	87.53	&	0.58	&	0.57	&	0.35	&	0.40	&	0.66	\\ \hline
3	&	98.66	&	85.63	&	98.14	&	85.69	&	86.25	&	0.34	&	0.32	&	0.38	&	0.43	&	0.64	\\ \hline
4	&	93.35	&	94.82	&	95.73	&	96.48	&	96.35	&	0.24	&	0.21	&	0.33	&	0.30	&	0.47	\\ \hline
5	&	99.30	&	99.16	&	97.96	&	87.73	&	97.18	&	0.29	&	0.17	&	0.19	&	0.36	&	0.68	\\ \hline
6	&	95.44	&	93.68	&	96.36	&	95.81	&	95.44	&	0.48	&	0.17	&	0.16	&	0.40	&	0.45	\\ \hline
7	&	98.69	&	98.92	&	99.15	&	98.36	&	98.12	&	0.14	&	0.15	&	0.14	&	0.27	&	0.37	\\ \hline
8	&	99.53	&	99.88	&	98.49	&	97.88	&	98.69	&	0.05	&	0.13	&	0.24	&	0.33	&	0.43	\\ \hline
9	&	96.41	&	96.53	&	95.67	&	89.50	&	73.12	&	0.10	&	0.12	&	0.32	&	0.45	&	0.32	\\ \hline
10	&	99.99	&	99.99	&	99.78	&	99.94	&	99.94	&	0.15	&	0.10	&	0.13	&	0.39	&	0.26	\\ \hline
\hline																			
Average	&	96.70	&	95.53	&	96.48	&	93.17	&	91.94	&	0.30	&	0.26	&	0.27	&	0.38	&	0.51	\\ \hline
\end{tabular}
\end{table}

\begin{table}[]
\small
\caption{Degree of Fuzziness (w) in Fuzzy-CNF for St. Vincent's  Dataset}
\label{tab:lw}
\noindent\hrulefill

\smallskip\noindent

\resizebox{\columnwidth}{!}{%
\begin{tabular}{|c|c|c|c|c|c|c|c|c|c|c|}
\multirow{2}{*}{Fold} & \multicolumn{5}{c|}{Accuracy (\%)} & %
    \multicolumn{5}{c|}{Computation Time (hours)}\\
\cline{2-11}
 & w=1.05 &  w=1.1  & w=1.2 & w=1.3  &  w=1.4  & w=1.05 &  w=1.1  & w=1.2 & w=1.3  &  w=1.4 \\
\hline
\hline
1	&	96.08	&	95.26	&	95.99	&	91.31	&	86.82	&	0.59	&	0.61	&	0.49	&	0.43	&	0.86	\\ \hline
2	&	89.56	&	91.41	&	87.49	&	88.96	&	87.53	&	0.58	&	0.57	&	0.35	&	0.40	&	0.66	\\ \hline
3	&	98.66	&	85.63	&	98.14	&	85.69	&	86.25	&	0.34	&	0.32	&	0.38	&	0.43	&	0.64	\\ \hline
4	&	93.35	&	94.82	&	95.73	&	96.48	&	96.35	&	0.24	&	0.21	&	0.33	&	0.30	&	0.47	\\ \hline
5	&	99.30	&	99.16	&	97.96	&	87.73	&	97.18	&	0.29	&	0.17	&	0.19	&	0.36	&	0.68	\\ \hline
6	&	95.44	&	93.68	&	96.36	&	95.81	&	95.44	&	0.48	&	0.17	&	0.16	&	0.40	&	0.45	\\ \hline
7	&	98.69	&	98.92	&	99.15	&	98.36	&	98.12	&	0.14	&	0.15	&	0.14	&	0.27	&	0.37	\\ \hline
8	&	99.53	&	99.88	&	98.49	&	97.88	&	98.69	&	0.05	&	0.13	&	0.24	&	0.33	&	0.43	\\ \hline
9	&	96.41	&	96.53	&	95.67	&	89.50	&	73.12	&	0.10	&	0.12	&	0.32	&	0.45	&	0.32	\\ \hline
10	&	99.99	&	99.99	&	99.78	&	99.94	&	99.94	&	0.15	&	0.10	&	0.13	&	0.39	&	0.26	\\ \hline
\hline																			
Average	&	96.70	&	95.53	&	96.48	&	93.17	&	91.94	&	0.30	&	0.26	&	0.27	&	0.38	&	0.51	\\ \hline
\end{tabular}
}
\end{table}

In implementing the FCM clustering to find a subset of features, we should also pay attention to the degree of fuzziness ($w$) is used. The higher $w$, a more blurred membership of a cluster. Hence, if the fuzziness is too high, then all objects have the same degree to all clusters. Therefore, it is necessary to find an optimal testing $w$ St. Vincent's dataset. The results are shown in Table  \ref{tab:lw}. The accuracy tendency is decreasing with greater the degree of fuzziness ($w$) except implementations with $w$=1.2. The computation time has an upward trend in line with the increasing degree of fuzziness except for $w$ = 1.05 which is higher than $w$ = 1.1.

\begin{table}[]
\small
\caption{Comparing all methods for St. Vincent's  Dataset}
\label{tab:lall}
\begin{tabular}{|c|c|c|}
\hline
 Methods & Accuracy (\%) & Computation Time (hours)  \\
\hline
\hline
CRF	&	94.34	&	0.58	\\ \hline
HCRF	&	86.64	&	1.96	\\ \hline
LDCRF	&	90.28	&	8.12	\\ \hline
CNF	&	95.98	&	0.10	\\ \hline
LDCNF	&	96.19	&	0.52	\\ \hline
DBN	&	84.47	&	0.04	\\ \hline
DBN - HMM	&	93.11	&	0.12	\\ \hline
DBN - CRF	&	93.51	&	0.35	\\ \hline
DBN - HCRF	&	89.97	&	0.95	\\ \hline
DBN - LDCRF	&	92.37	&	5.38	\\ \hline
DBN - CNF	&	96.58	&	0.12	\\ \hline
DBN - LDCNF	&	96.45	&	0.40	\\ \hline
FCM - CRF	&	92.89	&	0.36	\\ \hline
FCM - CNF	&	96.75	&	0.33	\\ \hline

\end{tabular}
\end{table}

The accuracy of CNF for this sleep stage classification ranges 91.07\% up to 95.80 \%. The performance of CNF increased when DBN and FCM applied. By using DBN, the achieved accuracy ranges 96.26 \% up to 96.58. Even this performance has the highest performance when FCM is implemented. It ranges 96.47 \% up to 96.75 \%. Thus, it can be seen that the two additional methods are proven effective in improving the accuracy of CNF.  Also, the CNF also proved as effective for classifying the sleep stage from St. Vincent's dataset with the accuracy above 90\%. An excellent performance of CNF for sleep stage classification can also be seen by comparing the other methods shown in Table  \ref{tab:lall}.

The lowest accuracy obtained when using DBN is 84.47 \%, then HCRF, LDCRF, CNF, and LDCNF although when they did not add any feature extraction. The HCRF gives lower accuracy than the CRF might be because it applied its intermediate layer but ignoring the factor of label interaction. When taking into account this label interaction factor, the LDCRF managed to give higher accuracy than HCRF. However, the accuracy of the LDCRF is still lower than the CRF. Maybe, it can be inferred that the addition of an intermediate layer causes a decrease in accuracy for sleep stage classification. On the other hand, the LDCNF accuracy is higher than the CNF. It means the gate function successfully find a subset of the features that have interacting features. By using this feature subset, the performance of the classification is increased when LDCNF also implements its hidden states to get the probabilistic models.

When applied as a classifier, DBN showed the lowest accuracy. When applied as feature extractor, however,  the DBN was proven effective in improving the accuracy of CRF, HCRF, LDCRF, CNF, and LDCNF. With DBN as feature extractor, the CNF has the highest accuracy than other methods which also use the DBN. From Table  \ref{tab:lall}, it also known that DBN combined with CRF, CNF, and LDCNF have the higher accuracy than the DBN on Hidden Markov Models which proposed by L\"{a}ngkvist et.al. 2012 on the same dataset. Moreover, the accuracy of CNF was able to achieve 96.58 \%. Also, by using FCM Clustering, the accuracy of CNF  increases of 96.75 \% but CRF decreased when compared using the DBN.

As for the computation time, the use of hidden state takes longer time than without hidden state. The HCRF Computation time was 1.96 hours, while for the CRF took only 0.58 hours when to apply the application of DBN as an additional feature extraction. The computation time increased when the external and internal structure of hidden states are both used. The LDCRF performance (with or without DBN as feature extractor). Besides that, it is also known that the gate function in CNF and LDCNF also proved effective to improve the accuracy and also decrease the computation time.

\subsection{Mitra Keluarga Kemayoran's  Dataset}

Mitra Keluarga Kemayoran's dataset also has five classes. To perform sleep stage classification at Mitra Keluarga Kemayoran's dataset, the DBN is used for feature extraction. The extracted feature subset consists of five labels that class. According to Table  \ref{tab:mg1}, the use of two to four gate gives accuracy within 96.99 \% to 98.9 \% range with a computation time ranges are from 0.82 to 0.89 hours for each fold. The accuracy of CNF decreased when FCM Clustering is used for feature extraction. The highest accuracy obtained is only 82.7 \% when using three gates. This implementation is described in detail in Table 8. From the Table, it can be seen that the accuracy of FCM Clustering did not increase with the addition of one or two more gates, on the other hand, the computation time was increased with the addition. This indicates that the use of FCM clustering is worse than the DBN for CNF classifier.

Tests in Table  \ref{tab:mg2} were performed to gain insight into the influence of the gate in CNF using a subset of the features derived from the FCM clustering. The test only focuses on the used number of gates, whereas the number of clusters and the degree of fuzziness ($w$) are not altered. The most optimal values for both two parameters are taken based on the Tables 9 and 10. In Table  \ref{tab:mcl}, the number of clusters is tested against its classification performance of the CNF. Tests carried out using four to eight clusters. From these tests, it is found that the optimal number of clusters is six. Increasing the number of clusters further did not improve the accuracy, but it might cost longer computation time. Whereas in  \ref{tab:mw}, it can be seen the influence of the degree of fuzziness ($w$) of the CNF. The result obtained optimal condition when $w = 1.05$.

Table  \ref{tab:mall} compared the CNF performance against CRF. It was known that CRF accuracy is lower than the CNF. However, when using DBN for feature extraction, the accuracy of CRF is higher than CNF, but the difference is only slight. However, unlike when FCM clustering was used feature extraction, the CRF accuracy is lower than the CNF. From this comparison, it is also learned that the gate function proved to be effective to reduce the computation time of the classification.

\begin{table}[]
\small
\caption{Number of Gates in DBN-CNF for Mitra Keluarga Kemayoran's  Dataset}
\label{tab:mg1}
\begin{tabular}{|c|c|c|c|c|c|c|}
\hline
\multirow{2}{*}{Fold} & \multicolumn{3}{c|}{Accuracy (\%)} & %
    \multicolumn{3}{c|}{Computation Time (hours)}\\
\cline{2-7}
 & g=2  &  g=3  & g=4  &  g=2  &  g=3  & g=4 \\
\hline
\hline
1	&	99.28	&	99.18	&	99.74	&	0.82	&	0.95	&	0.90	\\ \hline
2	&	95.41	&	98.47	&	99.31	&	0.79	&	0.83	&	0.87	\\ \hline
3	&	97.88	&	98.95	&	99.81	&	0.81	&	0.95	&	0.85	\\ \hline
4	&	99.77	&	99.87	&	99.91	&	0.92	&	0.95	&	0.88	\\ \hline
5	&	97.11	&	98.71	&	99.59	&	0.79	&	0.94	&	0.84	\\ \hline
6	&	95.96	&	98.67	&	99.78	&	0.87	&	0.80	&	0.91	\\ \hline
7	&	95.22	&	98.75	&	99.58	&	0.81	&	0.85	&	0.83	\\ \hline
8	&	93.69	&	98.65	&	99.63	&	0.80	&	0.84	&	0.89	\\ \hline
9	&	96.28	&	98.27	&	99.78	&	0.80	&	0.82	&	0.99	\\ \hline
10	&	99.31	&	99.49	&	99.77	&	0.79	&	0.81	&	0.89	\\ \hline
\hline													
Average	&	96.99	&	98.90	&	99.69	&	0.82	&	0.87	&	0.89	\\ \hline
\end{tabular}
\end{table}

\begin{table}[]
\small
\caption{Number of Gates in FCM-CNF for Mitra Keluarga Kemayoran's  Dataset}
\label{tab:mg2}
\begin{tabular}{|c|c|c|c|c|c|c|c|c|c|c|}
\hline
\multirow{2}{*}{Fold} & \multicolumn{4}{c|}{Accuracy (\%)} & %
    \multicolumn{4}{c|}{Computation Time (hours)}\\
\cline{2-9}
 & g=2  &  g=3  & g=4  &  g=5 & g=2  &  g=3  & g=4  &  g=5 \\
\hline
\hline
1	&	94.57	&	95.51	&	94.92	&	94.94	&	0.28	&	0.30	&	0.29	&	0.32	\\ \hline
2	&	76.64	&	76.74	&	77.64	&	76.19	&	0.33	&	0.46	&	0.37	&	0.37	\\ \hline
3	&	85.38	&	86.08	&	85.18	&	85.21	&	0.39	&	0.37	&	0.41	&	0.45	\\ \hline
4	&	96.49	&	96.61	&	95.72	&	95.76	&	0.28	&	0.32	&	0.28	&	0.31	\\ \hline
5	&	76.79	&	76.78	&	79.46	&	80.30	&	0.39	&	0.36	&	0.37	&	0.38	\\ \hline
6	&	69.96	&	72.19	&	57.54	&	57.58	&	0.25	&	0.34	&	0.24	&	0.31	\\ \hline
7	&	93.36	&	93.84	&	91.53	&	91.49	&	0.47	&	0.31	&	0.53	&	0.50	\\ \hline
8	&	44.67	&	48.40	&	46.59	&	47.39	&	0.37	&	0.25	&	0.30	&	0.33	\\ \hline
9	&	88.37	&	88.69	&	87.34	&	87.62	&	0.25	&	0.32	&	0.29	&	0.29	\\ \hline
10	&	91.74	&	92.22	&	91.87	&	91.88	&	0.33	&	0.39	&	0.53	&	0.40	\\ \hline
\hline																	
Average	&	81.80	&	82.70	&	80.78	&	80.84	&	0.34	&	0.34	&	0.36	&	0.37	\\ \hline
\end{tabular}
\end{table}

\begin{table}[]
\small
\caption{Number of Clusters in FCM-CNF for Mitra Keluarga Kemayoran's  Dataset}
\label{tab:mcl}
\begin{tabular}{|c|c|c|c|c|c|c|c|c|c|c|}
\hline
\multirow{2}{*}{Fold} & \multicolumn{5}{c|}{Accuracy (\%)} & %
    \multicolumn{5}{c|}{Computation Time (hours)}\\
\cline{2-11}
 & cl=4  &  cl=5  & cl=6  & cl=7  &  cl=8  & cl=4  &  cl=5  & cl=6  & cl=7  &  cl=8 \\
\hline
\hline
1	&	95.96	&	95.05	&	95.51	&	95.63	&	95.47	&	0.25	&	0.21	&	0.30	&	0.46	&	0.29	\\ \hline
2	&	79.62	&	78.99	&	76.74	&	79.80	&	77.33	&	0.40	&	0.20	&	0.46	&	0.31	&	0.23	\\ \hline
3	&	85.95	&	86.60	&	86.08	&	84.80	&	85.23	&	0.30	&	0.23	&	0.37	&	0.24	&	0.42	\\ \hline
4	&	92.60	&	96.67	&	96.61	&	96.68	&	95.65	&	0.31	&	0.22	&	0.32	&	0.27	&	0.53	\\ \hline
5	&	83.40	&	81.32	&	76.78	&	78.97	&	80.55	&	0.20	&	0.30	&	0.36	&	0.61	&	0.29	\\ \hline
6	&	53.54	&	71.88	&	72.19	&	74.80	&	72.36	&	0.25	&	0.23	&	0.34	&	0.27	&	0.19	\\ \hline
7	&	72.82	&	75.50	&	93.84	&	89.39	&	91.10	&	0.27	&	0.33	&	0.31	&	0.29	&	0.36	\\ \hline
8	&	52.22	&	44.57	&	48.40	&	44.49	&	44.60	&	0.30	&	0.24	&	0.25	&	0.54	&	0.58	\\ \hline
9	&	73.46	&	81.67	&	88.69	&	88.11	&	88.48	&	0.37	&	0.34	&	0.32	&	0.31	&	0.31	\\ \hline
10	&	91.57	&	91.39	&	92.22	&	90.79	&	91.70	&	0.18	&	0.30	&	0.39	&	0.21	&	0.63	\\ \hline
\hline																					
Average	&	78.11	&	80.36	&	82.70	&	82.35	&	82.25	&	0.28	&	0.26	&	0.34	&	0.35	&	0.38	\\ \hline

\end{tabular}
\end{table}

\begin{table}[]
\small
\caption{Degree of Fuzziness (w) in FCM-CNF for Mitra Keluarga Kemayoran's  Dataset}
\label{tab:mw}
\noindent\hrulefill

\smallskip\noindent

\resizebox{\columnwidth}{!}{%
\begin{tabular}{|c|c|c|c|c|c|c|c|c|c|c|}
\multirow{2}{*}{Fold} & \multicolumn{5}{c|}{Accuracy (\%)} & %
    \multicolumn{5}{c|}{Computation Time (hours)}\\
\cline{2-11}
 & w=1.05 &  w=1.1  & w=1.2 & w=1.3  &  w=1.4  & w=1.05 &  w=1.1  & w=1.2 & w=1.3  &  w=1.4 \\
\hline
\hline
1	&	95.51	&	95.25	&	95.46	&	94.17	&	93.71	&	0.30	&	0.51	&	0.36	&	0.43	&	0.46	\\ \hline
2	&	76.74	&	79.65	&	78.28	&	72.40	&	65.08	&	0.46	&	0.29	&	0.32	&	0.45	&	0.56	\\ \hline
3	&	86.08	&	85.23	&	86.83	&	85.92	&	85.52	&	0.37	&	0.52	&	0.87	&	0.72	&	0.58	\\ \hline
4	&	96.61	&	96.01	&	94.53	&	95.23	&	95.61	&	0.32	&	0.53	&	0.40	&	0.35	&	0.60	\\ \hline
5	&	76.78	&	79.55	&	79.42	&	77.23	&	76.93	&	0.36	&	0.49	&	0.42	&	0.36	&	0.41	\\ \hline
6	&	72.19	&	69.81	&	72.68	&	73.49	&	56.06	&	0.34	&	0.43	&	0.47	&	0.29	&	0.54	\\ \hline
7	&	93.84	&	92.80	&	92.64	&	88.37	&	88.49	&	0.31	&	0.24	&	0.25	&	0.49	&	0.39	\\ \hline
8	&	48.40	&	44.93	&	48.92	&	46.64	&	56.64	&	0.25	&	0.45	&	0.49	&	0.48	&	0.47	\\ \hline
9	&	88.69	&	88.90	&	86.86	&	74.24	&	73.61	&	0.32	&	0.46	&	0.36	&	0.66	&	0.65	\\ \hline
10	&	92.22	&	91.76	&	90.98	&	89.85	&	90.94	&	0.39	&	0.64	&	0.28	&	0.52	&	0.50	\\ \hline
\hline																				Average	&	82.70	&	82.39	&	82.66	&	79.75	&	78.26	&	0.34	&	0.46	&	0.42	&	0.47	&	0.52	\\ \hline

\end{tabular}
}
\end{table}

\begin{table}[]
\small
\caption{Comparing all methods for Mitra Keluarga Kemayoran's  Dataset}
\label{tab:mall}
\begin{tabular}{|c|c|c|}
\hline
  Methods & Accuracy (\%) & Computation Time (hours)  \\
\hline
\hline
CRF	&	75.59	&	0.58	\\ \hline
CNF	&	79.39	&	0.2	\\ \hline
DBN - CRF	&	99.96	&	1.02	\\ \hline
DBN - CNF	&	99.69	&	0.89	\\ \hline
FCM - CRF	&	76	&	0.46	\\ \hline
FCM - CNF	&	82.7	&	0.34	\\ \hline
\end{tabular}
\end{table}

\section{Conclusion}
This paper proposed the sequence-based treatment of sleep signals for sleep stage classification using Conditional Neural Fields (CNF). In this study, we compared the DBN and FCM clustering as feature extractors. The result showed that the CNF is mostly superior in accuracy and computation time compared with CRF, HCRF, LDCRF, and LDCNF. As for the feature extraction, On the other hand, we found that the DBN is better than the FCM clustering. In the first dataset, the accuracy of DBN is lower than the FCM clustering but with a very thin margin. While in the second dataset, the accuracy of from DBN feature extractor is much higher than the FCM Clustering. Since the FCM Clustering shown as effective methods for sleep stage classification in St. Vincent's Data, for our future study, it might be interesting to elaborate the use of another type of Fuzzy clustering, such as Fuzzy Subtractive Clustering (FSC). Unlike FCM, the FSC adaptively determined the number of effective clusters directly based on the data and membership computation based on density.

\section{Acknowledgment}
This work is supported by Higher Education Center of Excellence Research Grant funded Indonesia Ministry of Research and Higher Education Contract No. 1068/UN2.R12/ HKP.05.00/2016

\section{Conflict of Interests}
The authors declare that there is no conflict of interest regarding the publication of this paper.


\bibliography{library}

\bibliographystyle{abbrv}

\end{document}